\documentclass[10pt,twocolumn,letterpaper]{article}

\usepackage{cvpr}
\usepackage{times}
\usepackage{epsfig}
\usepackage{graphicx}
\usepackage{amsmath}
\usepackage{amssymb}
\usepackage{authblk}
\usepackage[normalem]{ulem}

\usepackage[pagebackref=true,breaklinks=true,letterpaper=true,colorlinks,bookmarks=false]{hyperref}

\cvprfinalcopy 

\begin{document} 

\title{Fashion Outfit Complementary Item Retrieval}

\author{Yen-Liang Lin, Son Tran, Larry S. Davis\\
Amazon\\
{\tt\small \{yenliang,sontran,lrrydav\}@amazon.com}
}

\maketitle

\begin{abstract}
Complementary fashion item recommendation is critical for fashion outfit completion. 
Existing methods mainly focus on outfit compatibility prediction but not in a retrieval setting.
We propose a new framework for outfit complementary item retrieval.
Specifically, a category-based subspace attention network is presented, which is a scalable approach for learning the subspace attentions. 
In addition, we introduce an outfit ranking loss that better models the item relationships of an entire outfit. 
We evaluate our method on the outfit compatibility, FITB and new retrieval tasks. 
Experimental results demonstrate that our approach outperforms state-of-the-art methods in both compatibility prediction and complementary item
retrieval.
\end{abstract}

\section{Introduction}
\begin{figure}[t]
\centering
\includegraphics[width=0.5\textwidth]{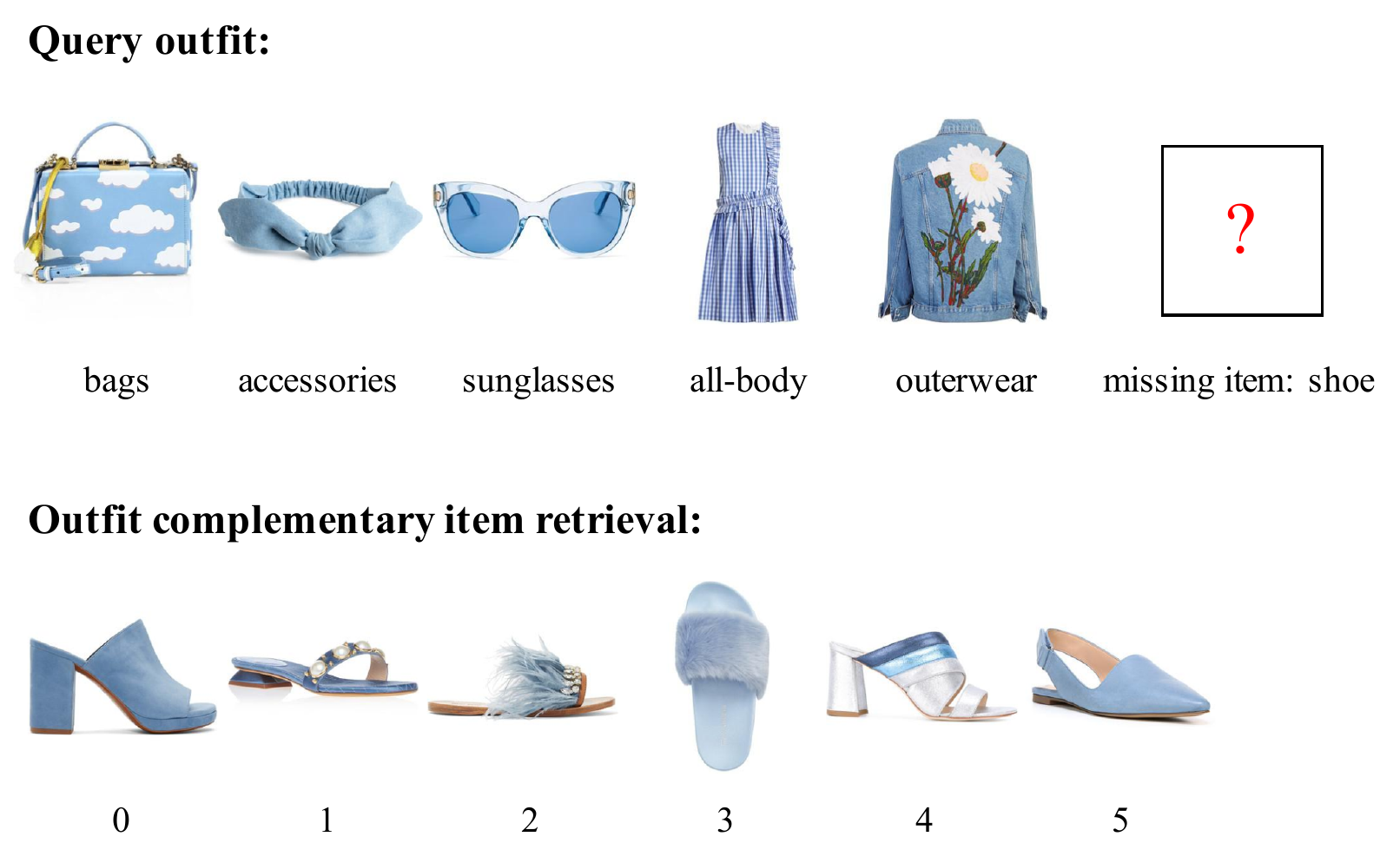}
\caption{An example of outfit complementary item retrieval. First row: partial outfit with a missing item (shoe). Second row: retrieved results using our method.}
\label{fig:figure1}
\end{figure}

Outfit complementary item retrieval is the task of finding compatible item(s) to complete an outfit. 
For example, given a (partially constructed) outfit with a top, a bottom and a pair of shoes, finding a bag that can go well (i.e., is compatible) with them. 
It is an important recommendation problem especially for online retailers. 
Customers frequently shop for clothing items that fit well with what has been selected or purchased before. 
Being able to recommend compatible items at the right moment would improve their shopping experience. 
Figure \ref{fig:figure1} gives an illustration of the problem.

Unlike visual or text similarity searching (e.g., \cite{zheng2019hardness, wang2019multi}), the item of interest (a.k.a. missing item or target item) for complementary item retrieval is from a different category than the categories in the incomplete outfit. 
So they are visually dis-similar from the other items in the outfit. 
They are instead consistent with the global outfit style. As discussed in \cite{Type-aware, SCE-net}, complementary fashion items are compatible along multiple attribute dimensions such as color, pattern, material, occasion. 
Being able to deal with this variety is essential. 
Moreover, a practical outfit completion system must support large scale searching,
where exhaustive comparison is prohibitive.

Prior work such as \cite{SiameseNet, mcauley2015image} addressed pairwise complementary retrieval but did not explicitly deal with entire outfit compatibility. 
Recent approaches such as \cite{MarylandPolyvore, SCE-net, Context-aware} focus on outfit compatibility prediction but not in a retrieval setting.
While technically it is possible to use the classification score to rank items in the database for retrieval, it is impractical to do so at a large scale. 
These approaches were not designed with large-scale outfit complementary item retrieval in mind (see Section \ref{sec:related_work} and Section \ref{sec:Experiment} for further analysis and comparison).

In this work, to capture the notion of item compatibility, we employ an approach that uses multiple style sub-spaces, similar to \cite{CSN, Type-aware, SCE-net}. 
In contrast to prior work, while the subspaces are generated using a shared model, the subspace attention weights are conditioned on the item categories. 
Our system is designed for large scale retrieval and outperforms the state-of-the-art on compatibility prediction, fill-in-the-blank (FITB) and outfit complementary item retrieval (see Section \ref{sec:Experiment}).

We train a deep neural network to generate feature embeddings for each item in an outfit. The network has three inputs: a source image, its category and a target category. The input image is passed through a convolutional neural network (CNN) for feature extraction. 
Two categorical signals are used to form an attention mechanism that biases the final embedding toward the area of the space that is supposedly for the source and target categories. 
We train this network in an end-to-end manner with a loss function that takes into account all outfit items. We evaluate our model on the typical FITB and outfit compatibility tasks (\cite{MarylandPolyvore, Type-aware, SCE-net, Context-aware}) using the published Polyvore Outfit dataset (\cite{Type-aware}). For the retrieval task, for each image in the database, feature extraction is carried out once for each of the target categories. Given the lack of benchmark dataset, we created a new one based on the existing Polyvore Outfit datasets and evaluate recall at top k of all approaches on these datasets. As shown in the experiment section, our approach outperforms state-of-the-art methods both in compatibility prediction and complementary item retrieval.

To summarize, the main technical contributions of our work are:
\begin{itemize}
\item We propose a category-based attention selection mechanism that only depends on the item categories. This enables us to conduct scalable indexing and searching for missing items in an outfit. 
\item We propose a new outfit ranking loss function that operates on entire outfits. This improves the accuracy in compatibility prediction as well as retrieval. 
\end{itemize} 

The rest of our paper is organized as follows. Section \ref{sec:related_work} reviews related work. 
Section \ref{sec:csa-net} and \ref{sec:outfit_ranking_loss} describe category-based subspace attention network and outfit ranking loss respectively. 
Section \ref{sec:retrieval} discusses the indexing and retrieval pipeline. 
Implementation details and experimental results are provided in Section \ref{sec:Experiment}. 
Section \ref{sec:conclusion} concludes the paper with a discussion about future steps.

\begin{figure*}[!t]
\centering
\includegraphics[width=1\textwidth]{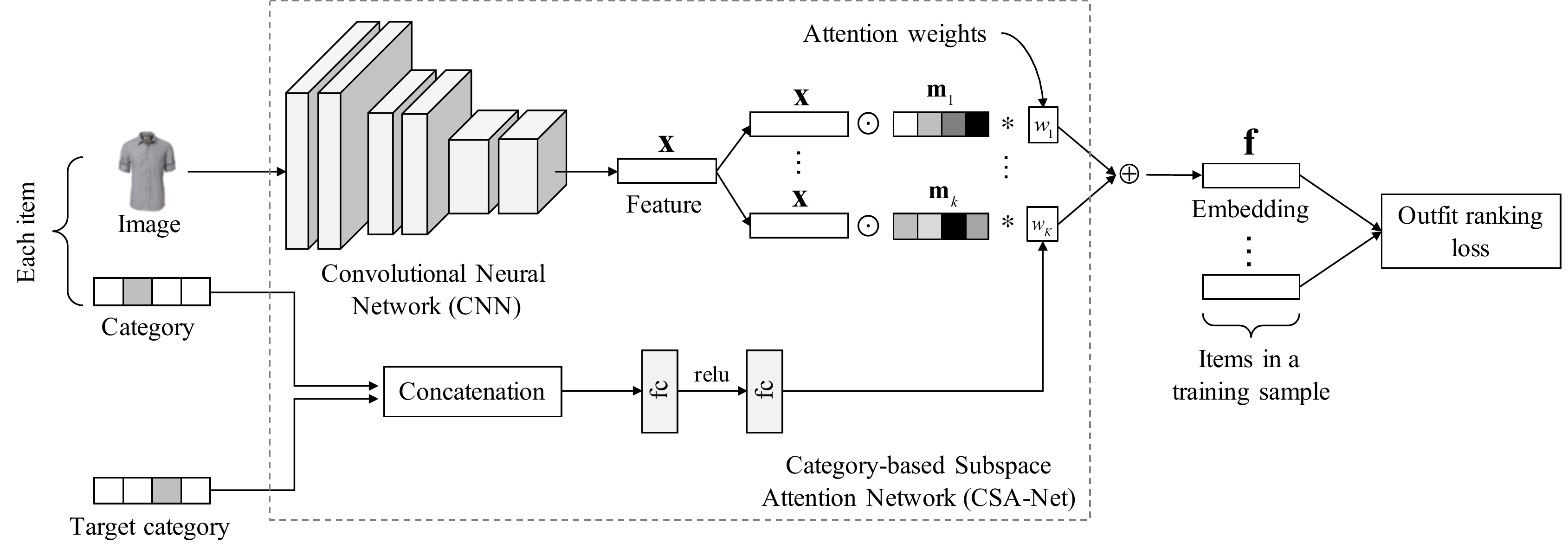}
\caption{Overview of our framework. Our framework takes a source image, its category vector and a target category vector as inputs.
The image is passed through a CNN to extract a visual feature vector, which is multiplied by a set of masks to obtain the subspace embeddings.
The concatenation of two category vectors is used to predict the subspace attention weights, which select the proper subspace embeddings for the final embedding computation.
%
%
Our network is trained by a ranking loss, which operates on the entire outfit.
%
}
\label{fig:system_overview}
\end{figure*}

\section{Related Work}
\label{sec:related_work}
\textbf{Outfit Compatibility Prediction:} 
Compatibility prediction has been addressed in a number of prior works at the pairwise item-to-item level (\cite{SiameseNet, mcauley2015image}) as well as at the entire outfit level (\cite{MarylandPolyvore, Type-aware, SCE-net}). 
In \cite{SiameseNet, mcauley2015image}, a common feature space (style space) was used for embedding items of different categories. 
The embedding model was learned using a large database of item co-purchase and co-view information. 
Item compatibility is determined by pairwise feature distances in the embedding space. 
The focus of these works was item-to-item, not entire outfits. 
Later works such as \cite{MarylandPolyvore, Type-aware, SCE-net, li2017mining} used outfit data from Polyvore and optimized for outfit-level compatibility. 
In \cite{MarylandPolyvore}, outfit items were treated as a sequences of input tokens to a bidirectional LSTM model. 
The LSTM make it possible to deal with outfits of varying size. This work also introduced the fill-in-the-blank task (FITB) which has been used in recent works such as (\cite{Context-aware, Type-aware, SCE-net}) for evaluation. 
For comparison purposes, we also benchmark our model on this task and the results are presented in Section \ref{sec:Experiment}. 
Graphical convolution network (GCN) \cite{GCN} was recently used in \cite{Context-aware} for outfit compatibility prediction. 
The graph was constructed by connecting items that occur together in outfits. 
The GCN was trained based on pairwise edge prediction. Highest performance is obtained by leveraging test-time graph connections, which is not practical, especially when new items can be introduced into the catalog (i.e., test set). New items added to a (dynamic) catalog often do not have any connections to existing items in the catalog. Therefore, it it unclear how to compute the embedding for the new items.
In general, these methods (e.g., \cite{MarylandPolyvore, Context-aware, SCE-net}) were designed and trained for compatibility classification, and were not optimized for retrieval.

Instead of computing the similarity in a single space, several recent approaches \cite{CSN, Type-aware, SCE-net} explored learning subspace embedding to capture different notions of similarities. 
Vasileva et al. \cite{Type-aware} adapted the CSN \cite{CSN} model to learn type-aware embeddings for modeling outfit compatibility. 
They learned a total of 66 conditional subspaces, each for a pair of item categories (e.g., tops-bottoms, tops-shoes, bottoms-shoes, etc). 
Tan et al. \cite{SCE-net} further improved the performance of \cite{Type-aware} by learning the shared subspaces as well as the importance of each subspace. 
Their method requires input image pairs during testing.
We also make use of shared subspace embedding, which helps to avoid dealing with a large number of models as in \cite{Type-aware}. 
Our subspace conditional factors are item categories, not target item images such as in \cite{SCE-net}, which enables us to carry out individual item feature extraction for large scale indexing and retrieval.

\textbf{Complementary Item Retrieval:}
Some of the compatibility prediction approaches mentioned above (\cite{SiameseNet, mcauley2015image, Type-aware})
can be used for retrieval. 
We compared our work with the most recent ones (\cite{Type-aware, SCE-net})
for retrieval on data extracted from existing Polyvore outfit datasets (\cite{MarylandPolyvore, Type-aware}).
Complementary item retrieval can be conducted using generative models such as generative adversarial neural network
(GAN), e.g., \cite{yu2019personalized, kang2017visually, shih2018compatibility}. 
Given an input image, a generative model is trained to generate a representation (e.g., image) of the target complementary item.
This generated item then can be used to retrieve real complementary items from an indexed database.
Most of these approaches address the pairwise retrieval task, and have been evaluated only on top-bottom or
bottom-to-top cases. 
It is not clear how well they will perform for other pairs of categories in a typical outfit, such as top-to-necklace or necklace-to-shoes, where fine details are critical. 
In recent work \cite{kang2019complete}, complementary items were retrieved based on the compatibility with scene images.
Global and local (scene patches) compatibilities measure the scene-product compatibility.
Instead, our approach focuses on the outfit complementary item retrieval, and considers the compatibility between the target image and every item in an outfit. 

\textbf{Distance Metric Learning:}
Distance metric learning is essential to retrieval. 
One can train a model using cross entropy loss (classification), hinge loss (e.g. \cite{weinberger2009distance}), triplet loss (\cite{schroff2015facenet, ge2018deep}), proxies (\cite{movshovitz2017no}), etc. In \cite{Type-aware, SCE-net}, the triplet loss was used but only for outfit compatibility prediction. 
In this work, we propose a new outfit ranking loss to better leverage the item relationships in an entire outfit.

\section{Proposed Approach}
Figure \ref{fig:system_overview} illustrates the system overview of our framework.
Our framework has three inputs: a source image, its category and a target category.
The input image is passed though a CNN to extract an image feature vector.
Similar to prior work \cite{CSN, SCE-net}, a set of masks are applied to the image feature vector to generate multiple subspace embeddings.
They correspond to different aspects of similarity. 
The concatenation of two categorical vectors are fed into a sub-network to predict the attention weights, whose purpose is to select the proper subspace embeddings for the final embedding computation.
The details of the category-based attention subspace network are presented in Section \ref{sec:csa-net}.

Our network is trained using a ranking loss that operates on the entire outfit. 
Given a training sample that includes an outfit $O$, a positive item $p$, and a set of negative items $N$, where 
the positive item is the one that is compatible to the outfit and negative items are ones that are not, 
the outfit ranking loss is computed based on the positive distance $d(O,p)$ and negative distance $d(O,N)$.
The whole framework is optimized in an end-to-end manner.
The details of the outfit ranking loss are presented in Section \ref{sec:outfit_ranking_loss}.

Our system is designed for large scale retrieval. 
We present our framework for indexing and outfit item retrieval in Section \ref{sec:retrieval}.

\subsection{Category-based Subspace Attention Networks}
\label{sec:csa-net}

In this section, we describe our category-based subspace attention network (CSA-Net).
Instead of computing the similarity in a single space, we utilize style subspaces (\cite{CSN, Type-aware, SCE-net}) to capture multiple dimensions of similarities.
This is important for complementary items, as they are visually dis-similar to other items in the outfit. 
Tan et al. \cite{SCE-net} learned shared subspaces, which achieved better performance than independent subspaces \cite{Type-aware}.
However, their method requires input image pairs for subspace selection during inference, which is not practical for retrieval. 
In contrast, our subspace attention mechanism only depends on the item categories.
Since our model only requires a single image and two category vectors, we can construct a practical indexing approach.
The network learns a non-linear function $\psi ({{\bf{I}}_s},{{\bf{c}}_s},{{\bf{c}}_t})$ that takes a source image ${{\bf{I}}_s}$, its category vector ${{\bf{c}}_s}$ and the target category vector ${{\bf{c}}_t}$ as inputs and generates a feature embedding ${\bf{f}}$.
The image is first passed into a CNN to extract its visual feature vector (denoted as ${\bf{x}}$).
The concatenation of two category vectors (one-hot encoding of the semantic category) is used to predict the subspace attention weights (${w_1} \ldots {w_k}$) ($k$ is the number of subspaces) using a sub-network, which contains two fully connected layers and a soft-max layer at the end.  
Then, a set of learnable masks (${{\bf{m}}_1},...,{{\bf{m}}_k}$) that have the same dimensionality as the image feature vector are applied to the image feature vector via Hadamard product.
These masks are learned so as to project the features into a set of subspaces that encode different similarity substructures. 
The final embedding constructed by the network is a weighted sum of the subspace embeddings:
\begin{equation}
	{\bf{f}} = \sum\limits_{i = 1}^k {({\bf{x}} \odot {{\bf{m}}_i})*{w_i}},
\end{equation}
where $k$ is the number of subspaces, ${\bf{x}}$ is the image feature vector after the base CNN, ${{\bf{m}}_i}$ is a learnable mask, ${w_i}$ is an attention weight, and ${\bf{f}}$ is the final embedding.

\begin{figure}[!t]
\centering
\includegraphics[width=0.5\textwidth]{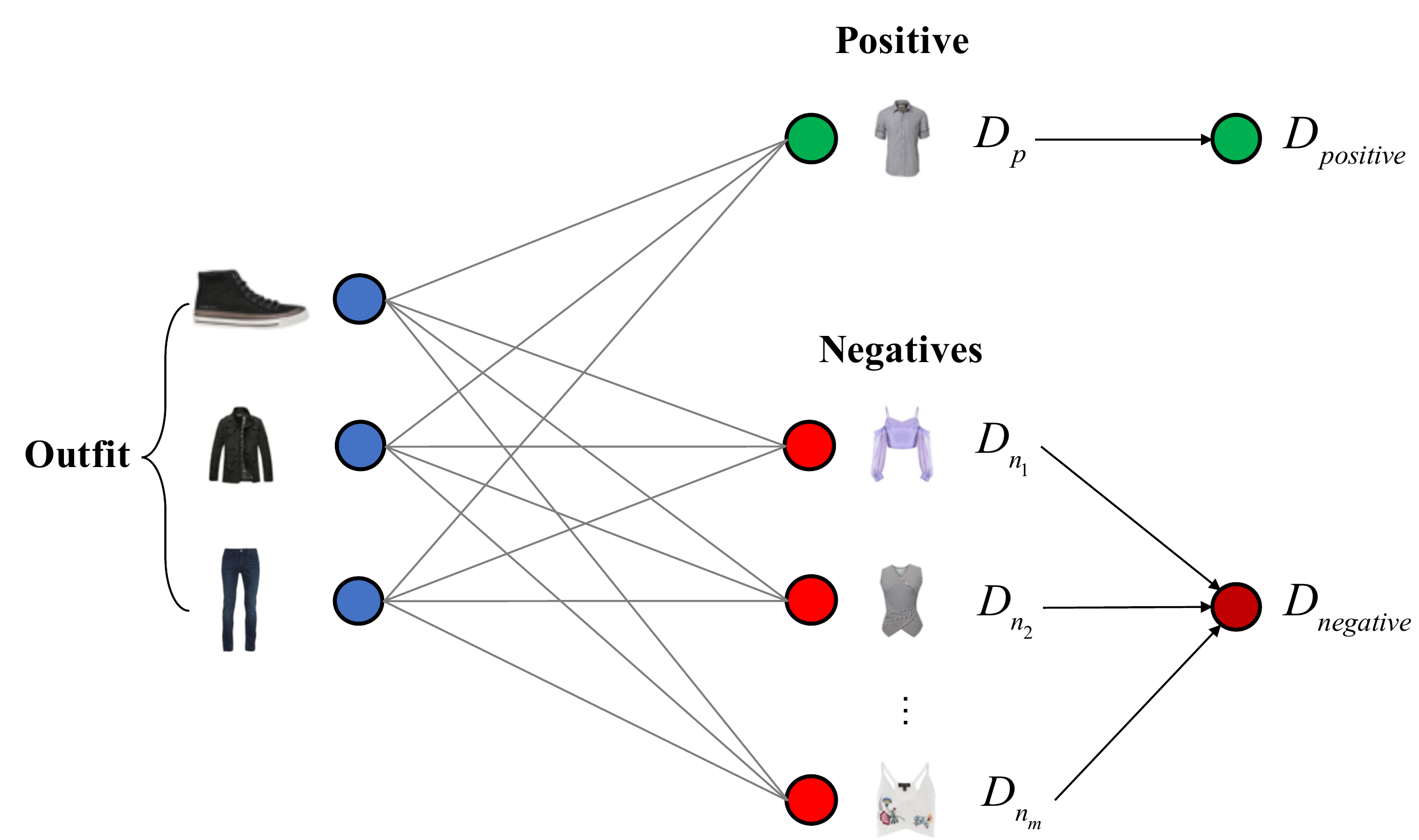}
\caption{Illustration of the proposed outfit ranking loss that considers the item relationship of the entire outfit.}
\label{fig:outfit_loss}
\end{figure}

\subsection{Outfit Ranking Loss}
\label{sec:outfit_ranking_loss}
Existing approaches \cite{Type-aware, SCE-net} use a triplet loss to learn feature embeddings for outfit compatibility prediction.
To form a triplet, a random pair of images (anchor and positive images) is first selected from an outfit, 
and a negative image that has the same semantic category as the positive image is randomly sampled. 
A triplet loss is optimized by pushing the negative sample away from the anchor image by a distance greater than a margin compared to the positive sample.
In contrast, when determining the compatibility of a complementary item to an outfit, we consider the similarities of all existing items in the outfit, not just a single item.

Figure \ref{fig:outfit_loss} illustrates the outfit ranking loss.
At each training iteration, we sample a mini-batch of triples, each of which $\Upsilon  = \{O,p,N\}$ consists of an outfit that contains a set of images $O = \{ {\bf{I}}_1^o,..,{\bf{I}}_n^o\}$, a positive image $p = \{ {\bf{I}}_{}^p\}$ that goes well with the outfit, and a set of negative images $N = \{ {\bf{I}}_1^n,...,{\bf{I}}_m^n\}$ that do not match the outfit.  
The feature embeddings of each image in the outfit $O$ are computed by:
\begin{equation}
	{\bf{f}}_i^o = \psi ({\bf{I}}_i^o,c({\bf{I}}_i^o),c({{\bf{I}}^p}));i = 1 \to n,
\end{equation}
where ${\bf{I}}_i^o$ is an image in $O$, $\psi ( \cdot )$ is our category-based subspace attention network, $c(.)$ is a function that maps an input image to an one-hot category vector, $c({\bf{I}}_i^o)$ is the source category vector for image ${\bf{I}}_i^o$ and $c({\bf{I}}_{}^p)$ is the target category vector from the positive image ${\bf{I}}_{}^p$.

Similarly, the feature embeddings for the positive image are computed by: 
\begin{equation}
	{\bf{f}}_i^p = \psi ({{\bf{I}}^p},c({\bf{I}}_i^o),c({{\bf{I}}^p}));i = 1 \to n
\end{equation}
A positive image has multiple embeddings ($i = 1 \to n$), as different category vectors $c({\bf{I}}_i^o)$ generate different subspace attentions, which results in different embeddings.
These embeddings will be used in pairwise distance computation (see Equation \ref{eq:outfit_dist}).
We also observe that the order of the input categories (e.g., $c({\bf{I}}_i^o)$ and $c({{\bf{I}}^p})$) does not affect the system performance; see discussions of order flipping in Section. \ref{sec:fitb_experiment}.

The feature embeddings for each negative image are computed by:
\begin{equation}
	{\bf{f}}_i^{{n_j}} = \psi ({\bf{I}}_j^n,c({\bf{I}}_i^o),c({\bf{I}}_j^n)),j = 1 \to m,i = 1 \to n
\end{equation}
Similar to the positive image, each negative image has multiple embeddings using different category vectors $c({\bf{I}}_i^o)$.

We define the distance of each item $s$ (positive or negative item) to the entire outfit by:
\begin{equation}
	{D_{outfit}}(O,s) = \frac{1}{n}\sum\limits_{i = 1}^n {d({\bf{f}}_i^o,{\bf{f}}_i^s)},
\label{eq:outfit_dist}
\end{equation}
where ${i}$ is an image in the outfit $O$, ${\bf{f}}_i^o$ and ${\bf{f}}_i^s$ are the feature embeddings for image ${i}$ and $s$ respectively, and ${d({\bf{f}}_i^o,{\bf{f}}_i^s)}$ is the pairwise distance between two images. 
Using Equation \ref{eq:outfit_dist}, we can compute the distance of the positive item ${D_p} = {D_{outfit}}(O,p)$ and negative items ${D_{{n_j}}} = {D_{outfit}}(O,{n_j})$ for $j = 1$ to $m$.
We then combine the ${D_{{n_i}}}$ using an aggregation function $\varphi$ (e.g., min or average) that aggregates all the negative items from the negative set into a single distance: 
\begin{equation}
	{D_N} = \varphi ({D_{{n_1}}},...,{D_{{n_m}}})
\end{equation}
This outfit ranking loss encourages the network to find an embedding where the distance between the outfit and negative samples is larger than the distance between the outfit and positive sample by a distance margin $m$:
\begin{equation}
	l(O,p,N) = \max (0,{D_{p}} - {D_{N}} + m),
\end{equation}
where ${D_p}$ and ${D_N}$ are the positive and negative distance respectively.
In contrast to previous triplet losses, our outfit ranking loss computes distances based on the entire outfit rather than a single item \cite{Type-aware, SCE-net}. 

\begin{figure}[!t]
\centering
\includegraphics[width=0.5\textwidth]{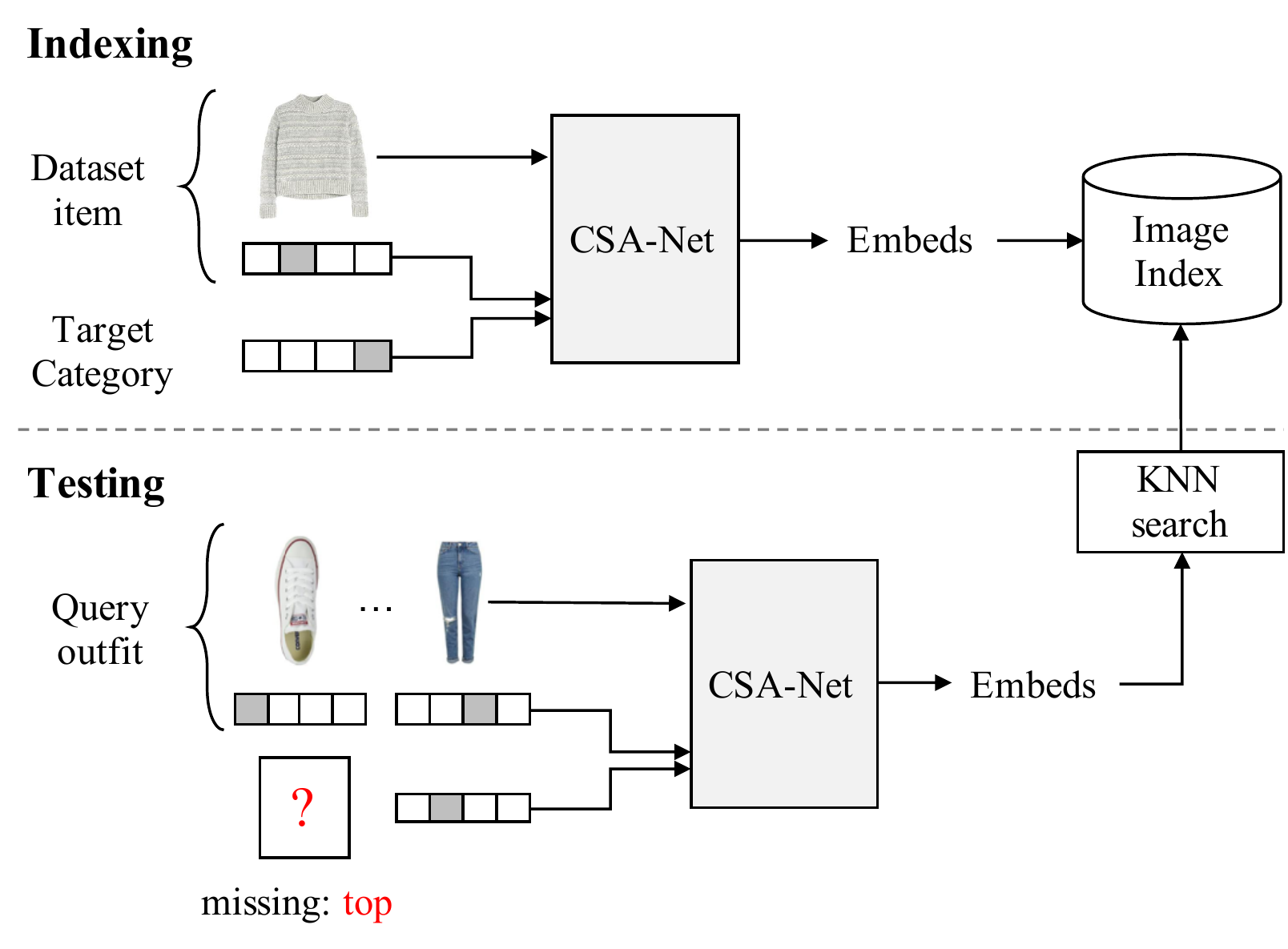}
\caption{Our framework for outfit complementary item retrieval. 
For indexing, multiple embeddings of each item are computed by enumerating the target categories.
More detailed explanations of category enumeration are in Section. \ref{sec:retrieval}.
For testing, given a query item in an outfit, we compute its embedding by using its image, category vector and the target category vector into our model. 
KNN search is used to retrieve compatible items from the indexed dataset.
The final ranking is obtained by fusing the scores from different query items.}
\label{fig:retrieval}
\end{figure}

\subsection{Outfit Complementary Item Retrieval}
\label{sec:retrieval}
Outfit complementary item retrieval is the task of retrieving a set of compatible items that cohesively match the existing items in an outfit.
Efficient retrieval involves feature extraction and indexing, which avoids linearly scanning the entire dataset during testing.
Given a query image (outfit) and a target category, the system extracts its embeddings and uses it to query the indexed dataset images.

Previous approaches for compatibility prediction are not suitable for retrieval.
\cite{SCE-net} utilizes pairs of images, which requires exhaustive pairwise comparison during testing and makes it impractical for indexing (it is impractical to generate the target (query) images during indexing). 
%
%
%
Cucurull et al. \cite{Context-aware} utilize graphical convolution neural (GCN) networks for outfit compatibility. However, it is unclear how to adapt their classification model to retrieval.

In contrast, we propose a new outfit complementary item retrieval system based on our category-based subspace attention network.
Figure \ref{fig:retrieval} illustrates our framework for retrieval.  
For feature extraction, multiple embeddings of each item are computed by enumerating the target categories.
This is because that different category pairings generate different subspace attentions, which result in different embeddings.
Query images can be from any category during testing; therefore we need to compute the embeddings from different target categories during indexing.
For example, the category vector of a shoe item is concatenated with top, bottom, etc category vectors to generate multiple embeddings during indexing. 
Given a top item during testing, we then search the shoe embeddings that are pre-computed by the top category.     
Images from each category are paired with different target categories, therefore the embedding size is linear in the number of high-level categories (e.g., 11 semantic categories in Polyvore-Outfit dataset) 
We use off-the-shelf approximate nearest neighbor search packages (e.g., \cite{Hnswlib}) for indexing and retrieval.

During testing, given an outfit, we retrieve a set of compatible images for the target category. 
For each item in the query outfit, we use its image, category vector and the target category vector in our category-based subspace attention network to extract its embedding.
Then, we use the embedding to search compatible items in the dataset though k-nearest neighbor (KNN) search.
We perform similar steps for each item in the query outfit and use an aggregation function (e.g., average fusion) to fuse the ranking scores from different query items to obtain the final rankings. 

\begin{table*}[]
\center
\small
\begin{tabular}{|l|l|c|c|c|c|}
\hline
                                     			&                 		& \multicolumn{2}{c|}{Polyvore Outfits-D}                               & \multicolumn{2}{c|}{Polyvore Outfits}                                 \\ \hline
Method                               		& Feature          	& \multicolumn{1}{l|}{FITB Accuracy} & \multicolumn{1}{l|}{Compat. AUC} & \multicolumn{1}{l|}{FITB Accuracy} & \multicolumn{1}{l|}{Compat. AUC} \\ \hline
Siamese-Net  \cite{Type-aware}        & ResNet18         	& 51.80                              & 0.81                             & 52.90                              & 0.81                             \\ \hline
Type-aware \cite{Type-aware}           & ResNet18 + Text  	& 55.65                              & 0.84                             & 57.83                              & 0.87                             \\ \hline
SCE-Net average  \cite{SCE-net}      & ResNet18 + Text 	& 53.67                              & 0.82                             & 59.07                              & 0.88                             \\ \hline \hline
CSA-Net + outfit ranking loss (ours) 	& ResNet18         	& \bf{59.26}                       	& \bf{0.87}                       & \bf{63.73}                        & \bf{0.91}                      \\ \hline
\end{tabular}
\caption{Comparison of different methods on the Polyvore-Outfit dataset (where -D denotes for disjoint set). We report the results of our method and state-of-the-art retrieval methods: Type-aware \cite{Type-aware} and SCE-Net average \cite{SCE-net} (Note that the original SCE-Net reports 61.6 FITB accuracy and 0.91 Compat. AUC on the non-disjoint set (Polyvore Outfits), which requires pairs of input images and is not designed for retrieval). All the methods use ResNet18 and embedding size 64 for fair comparison. Note that our method does not use text feature. Our category-based subspace attention network and outfit ranking loss shows superior performance compared to the baseline methods.}\label{table:fitb_and_compat}
\end{table*}

\section{Experiment}
\label{sec:Experiment}
We evaluate our method on the compatibility prediction, fill-in-the-blank (FITB) and outfit complementary item retrieval tasks.
\subsection{Evaluation}
\label{sec:dataset}
\textbf{Outfit compatibility and FITB:}
For compatibility prediction and FITB, we compare to state-of-the-art methods on the Polyvore Outfit \cite{Type-aware} dataset, which is the largest dataset for compatibility prediction.
The dataset includes disjoint and non-disjoint sets.
For the non-disjoint set, some items (but not complete outfits) may be seen in both training and test splits.
For the disjoint set, outfits in the test/validation set do not share any items in common with outfits in the training set.
The non-disjoint set is much larger than the disjoint set, which contains 53,306 training outfits and 10,000 testing outfits, while the disjoint set contains 16,995 training outfits and 15,145 testing outfits. 

The incorrect choices in each question of the FITB task in Polyvore Outfit dataset are sampled from the items that have the same category as the correct choice, 
while the FITB task in Maryland Polyvore \cite{MarylandPolyvore} are sampled at random without the category constraint.
The Polyvore Outfit dataset also provides fine-grained item type annotations. 

For outfit compatibility prediction, the task is to predict the compatibility of a set of fashion items in an outfit.
We use the standard metric AUC \cite{MarylandPolyvore} to evaluate the performance of compatibility prediction, which measures the area under the receiver operating characteristic curve.
For FITB, given a subset of items in an outfit and a set of candidate items (four items, one positive and three negatives), the task is to select the most compatible candidate.
The performance is evaluated based on the overall accuracy \cite{MarylandPolyvore}. 

\textbf{Outfit complementary item retrieval}: 
There is no ground truth retrieval annotations for the current datasets.
We created a new dataset based on Polyvore Outfit dataset for this purpose. 
We use the outfits in the FITB test set as the query outfits, and evaluate the effectiveness of an algorithm by measuring the rank of the positive image (recall@top k).
This provides a quantitative evaluation metric, so the results are reproducible and can be compared across different algorithms.

The rank of the positive image is not a perfect indicator of the utility of a system, since the database can contain many complementary items to the query outfits,
some of which may, in fact, be judged by human experts to be better matches to the outfit than the positive, but only one image is annotated as ground truth. 
The relative rank is still useful for comparing different algorithms, since better retrieval algorithms should have higher recall, as images that have similar styles as the positive image will move forward in the ranking.

We limit our image ranking results to be of the same fine-grained category as the positive image. 
Since the test set does not have enough fine-grained images for the retrieval experiments, 
we augment the testing images by including the training images as distractors, and select the fine-grained categories that have more than 3000 images. 
We use 3000 images for the categories that have more than 3000 images, so all categories have an equal number of images.
We totally select 27/153 fine-grained categories for the non-disjoint set, and 16/153 fine-grained categories for the disjoint set for the retrieval experiments. 

\subsection{Implementation Details}
\label{sec:implementation_details}
We train our category-based subspace attention network using the outfit ranking loss.
We use ResNet18 \cite{ResNet} as our backbone CNN model and embedding size 64 similar to state-of-the-art methods \cite{Type-aware, SCE-net} for fair comparison, which is pre-trained on Imagenet.
We set the number of subspaces to 5 as in \cite{SCE-net}.
We compute the ranking loss in an online mining manner similar to \cite{schroff2015facenet}, so the subspace embeddings of each image are only computed once and re-used for different pairwise distances. 
Specifically, given a training triple, we first pass all images though the CNN and category-based subspace attention network to extract subspace embeddings.
For each pair of items, we compute the subspace attention weights using two category vectors and multiply them by the pre-computed subspace embeddings to obtain the final embeddings.
For negative samples, because there is no ground truth negative images for each outfit, we randomly sample a set of negative images that have the same category as the positive image similar to \cite{Type-aware, SCE-net}.
We select semi-hard negative images to train the outfit ranking loss.    
We train the networks with a mini-batch size of 96 and optimize using ADAM. 
We set the margin to 0.3 for the ranking loss and initial learning rate to $5{e^{ - 5}}$.
We adopt a learning rate schedule that linearly decreases the learning rate to zero, but set the warmup ratio to zero as our initial learning rate is already small (see e.g., \cite{BERT}).

\subsection{Baselines}
We compare our method with several baseline approaches on the compatibility prediction, fill-in-the-blank (FITB) and outfit complementary item retrieval tasks.
As there is no existing approach for outfit complementary item retrieval, we modify the existing compatibility prediction methods \cite{Type-aware, SCE-net} for retrieval.

\textbf{Siamese-Net} \cite{Type-aware}: The approach of Veit et al. \cite{SiameseNet} that uses ResNet18 to learn the embeddings in a single space.

\textbf{Type-aware} \cite{Type-aware}:  We use the latest code from the authors and re-train the model on Polyvore Outfit dataset, which shows better performance than originally reported in the paper. For indexing, we project each image into 66 type subspaces to generate the features. During testing, for each testing item in an outfit, we use the types (from pairs of images) to retrieve the feature indexes and rank the images in that space. Average fusion is used to fuse scores from different items. 

\textbf{SCE-Net} \cite{SCE-net}: We use the code from the authors and re-train the model on Polyvore Outfit dataset. Their original model can not perform image retrieval, therefore we compare their average model as our baseline for the retrieval experiments. The average model computes the features by averaging the outputs from similarity condition masks. Average fusion is used to fuse scores from different items.

Note that Cucurull et al. \cite{Context-aware} utilize graph convolutional network (GCN) for fashion compatibility prediction. 
However, their method uses a classification model for compatibility prediction, which is not directly applicable to the retrieval task.  
%

\subsection{Compatibility and FITB Experiments}
\label{sec:fitb_experiment}
Though our method is designed for retrieval, we also report our results on the fill-in-the-blank (FITB) and outfit compatibility tasks (see Table \ref{table:fitb_and_compat}).
For SCE-Net average \cite{SCE-net}, the re-trained model achieves similar results for the non-disjoint set as reported in the original paper. 
As the authors do not report performance on the disjoint set, we report the performance based on our own run.
We also compare with a better version of type-aware method \cite{Type-aware}, the re-trained model achieves better performance than the original model.

We observe a pronounced improvement using our model over the SCE-Net average model, resulting in $4$ to $5\%$ improvement on FITB, and $3$ to $5\%$ improvement on outfit compatibility.
Notably, our result (63.73 FITB accuracy and 0.91 AUC in non-disjoint set) achieves even better performance than the original SCE-Net, which obtains the state-of-the-art performance (61.6 FITB accuracy and 0.91 AUC) on this dataset.
Also, our method does not use text features as the baseline methods \cite{Type-aware, SCE-net}, which use texts for learning a joint visual and text embedding.
We find that the performance is lower on the disjoint set for all methods, probably because the training set is much smaller than the non-disjoint set, and learning sub-space embeddings requires more training samples. 

\textbf{Triplet loss versus outfit ranking loss:}
We compare the performance between triplet loss and our outfit ranking loss (see Table \ref{table:loss_function}).
We evaluate on different settings where our model is trained on the original triplet loss versus outfit ranking loss.
The result shows that our outfit ranking loss further improves the performance of triplet loss, $\sim 3\%$ in the FITB task.  
%

\begin{table}[!htbp]
\small
\centering
\begin{tabular}{|l|c|c|}
\hline
Loss function					& FITB Accuracy 	& Compat. AUC 	\\ \hline
Triplet loss					& 56.17 / 60.91 	& 0.85 / 0.90	 	\\ \hline	
Outfit ranking loss      			& \bf{59.26 / 63.73}   & \bf{0.87 / 0.91}       \\ \hline					
\end{tabular}
\caption{Comparison of triplet loss and outfit ranking loss of our model in disjoint / non-disjoint set.}
\label{table:loss_function}
\end{table}

\textbf{Min versus average aggregation function:}
We compare different aggregation functions in our outfit ranking loss (see Table \ref{table:agg_function}). 
We observe that the min aggregation function achieves better performance than the average function, because it selects more hard negative examples into the ranking loss training. 

\begin{table}[!htbp]
\small
\centering
\begin{tabular}{|l|c|c|}
\hline
Aggregation function         & FITB Accuracy 	& Compat. AUC 	\\ \hline
Average 				& 56.19 / 60		& 0.84 / 0.89	 	\\ \hline	
Min                 			& \bf{59.26 / 63.73}   &\bf{0.87 / 0.91}       \\ \hline					
\end{tabular}
\caption{Comparison of different aggregation functions of our outfit ranking loss in disjoint / non-disjoint set.}
\label{table:agg_function}
\end{table}

\textbf{Order flipping:}
We also experimented with order flipping in our outfit ranking loss, where we flip the order of input categories when training our framework.
However, we did not see much improvement with flipping, probably because the network has seen enough pairs so the learned feature embedding generalizes to different orders.  


\begin{table*}[!t]
\center
\begin{tabular}{|l|l|l|l|l|l|l|}
\hline
                                    					& \multicolumn{3}{c|}{Polyvore Outfits-D}               		& \multicolumn{3}{c|}{Polyvore Outfits}                            						\\ \hline
Method                              					& Top 10          		& Top 30           		& Top 50          			& Top 10          		& Top 30           		& \multicolumn{1}{c|}{Top 50} 	\\ \hline
Type-aware \cite{Type-aware}           			& 3.66\%          		& 8.26\%           	& 11.98\%        	 		& 3.50\%          		& 8.56\%           	& 12.66\%                    		\\ \hline
SCE-Net average  \cite{SCE-net}     			& 4.41\%          		& 9.85\%           	& 13.87\%          		& 5.10\%          		& 11.20\%          	& 15.93\%                     		\\ \hline\hline
CSA-Net + outfit ranking loss (ours) 			& \textbf{5.93\%} 	& \textbf{12.31\%} 	& \textbf{17.85\%} 		& \textbf{8.27\%} 	& \textbf{15.67\%} 	& \textbf{20.91\%}   			\\ \hline
\end{tabular}
\caption{Comparison of different methods in complementary outfit item retrieval task (recall@top k). Our method shows a consistent improvement over the baseline approaches for different k.}
\label{table:retrieval}
\end{table*}

\subsection{Retrieval Experiments}
Table \ref{table:retrieval} shows the retrieval results, where we evaluate the performance using recall@top k (k=10, 30, 50) on 27/153 and 16/153 fine-grained categories in the non-disjoint and disjoint set respectively.  
Each fine-grained category has an equal number of images (e.g., 3000), so that the recall will not be biased to certain categories with more images. 
Our reported number is the mean of average recall for all categories.
The original SCE-Net \cite{SCE-net} can not perform retrieval as explained in Section \ref{sec:retrieval}, therefore we compare to their average model for the retrieval experiments. 

Our method shows a consistent improvement over the baseline approaches for different k. 
We obtain about $2$ to $5\%$ improvements over SCE-Net average model for recall@top 10, top 30 and top 50 respectively for both disjoint and non-disjoint set.
As explained in Section \ref{sec:dataset}, the metric may not reflect the practical performance of a system, as there may have many compatible items in the testing catalog. 
Some example retrieval results of our framework are shown in the Figure \ref{fig:result}. 

 
\begin{figure*}[!h]
\centering
\includegraphics[width=1.0\textwidth]{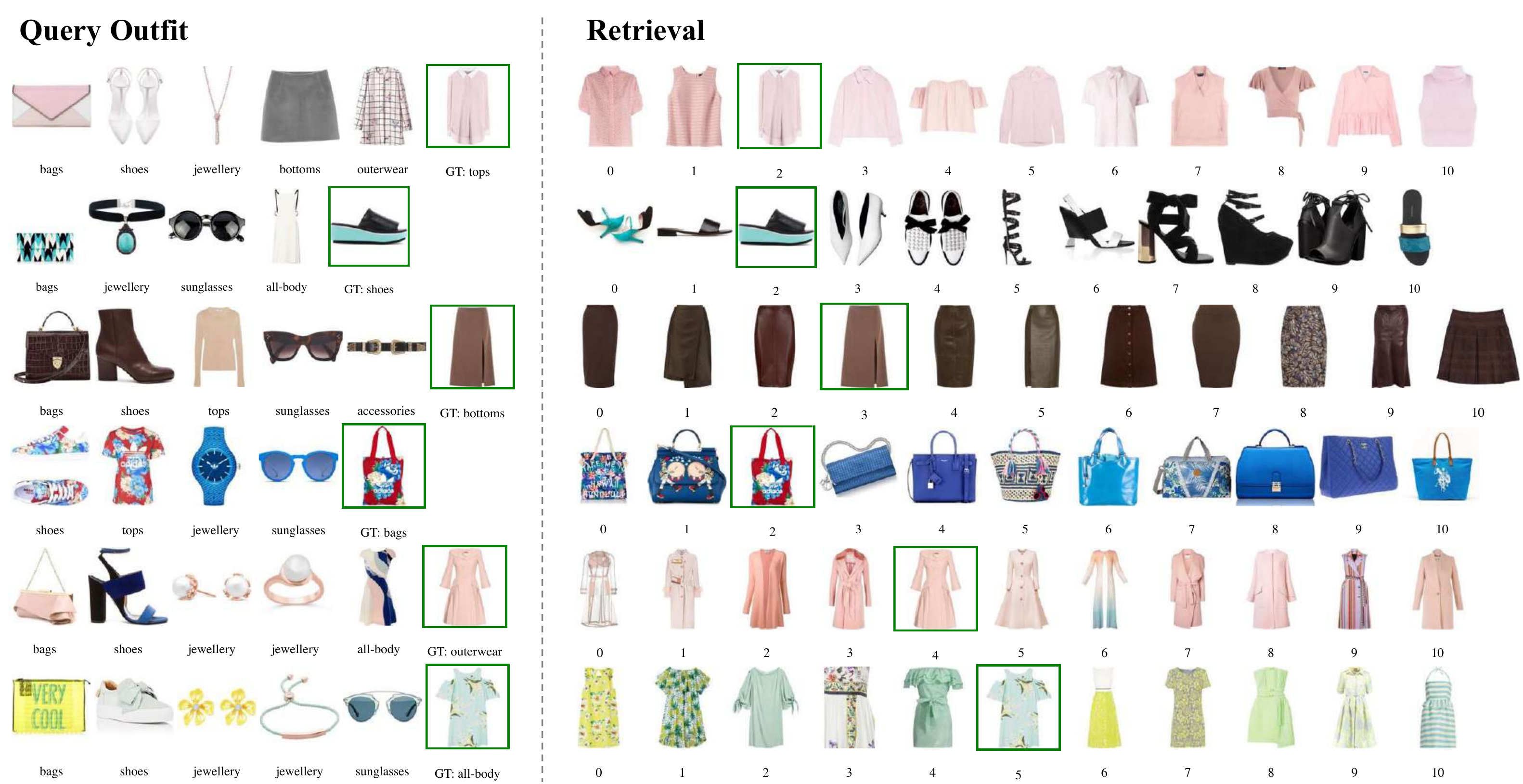}
\caption{Example retrieval results of our method. Left column: query (incomplete) outfits and (missing) target items. Right column: top 10 retrieval results over 3000 images, where ground truth items are in green boxes. Our approach retrieves a list of compatible items that match to the style of the entire outfit.}
\label{fig:result}
\end{figure*}
\begin{figure*}[!h]
\centering
\includegraphics[width=1.0\textwidth]{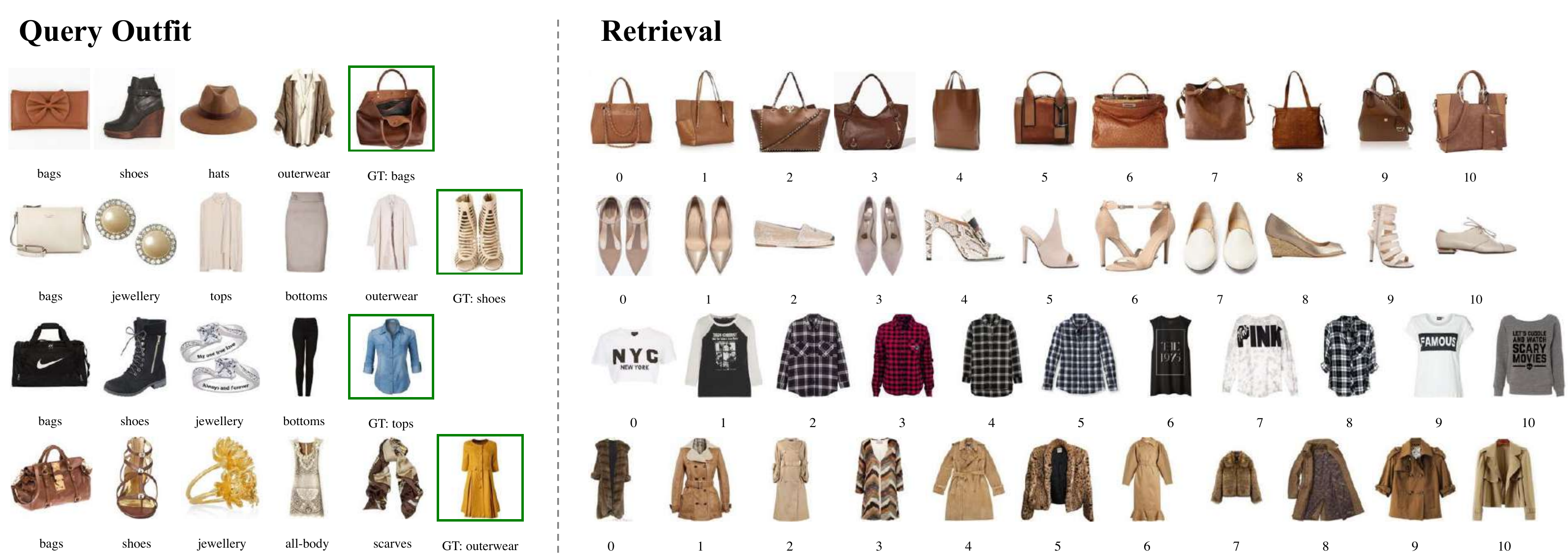}
\caption{Some failure cases of our method. The failure cases are mainly caused by similar colors, textures and styles to the target item (row 1 and row 2), and multiple compatible items (row 3 and row 4) that also match to the query outfit. Note that these items may not be necessarily incompatible with the query outfits, see Section \ref{sec:dataset} for the discussions about the dataset limitations.}
\label{fig:result}
\end{figure*}

\section{Conclusion}
\label{sec:conclusion}
We presented a new approach for outfit complementary item retrieval.
To overcome the indexing problems of existing approaches, 
we designed a category-based subspace attention network that takes a single image and two category vectors to form an attention mechanism, which makes our framework scalable for indexing and retrieval.  
In addition, we introduce an outfit ranking loss that considers the item relationship of an entire outfit to improve the outfit compatibility. 
The experimental results demonstrate that our model outperforms several state-of-the-art approaches in outfit compatibility, FITB and retrieval tasks.

{\small
\bibliographystyle{ieee_fullname}
\bibliography{egbib}

\begin{thebibliography}{10}\itemsep=-1pt

\bibitem{Hnswlib}
https://github.com/nmslib/hnswlib.

\bibitem{Context-aware}
Guillem Cucurull, Perouz Taslakian, and David Vazquez.
\newblock Context-aware visual compatibility prediction.
\newblock In {\em CVPR}, 2019.

\bibitem{BERT}
Jacob Devlin, Ming-Wei Chang, Kenton Lee, and Kristina Toutanova.
\newblock {BERT:} pre-training of deep bidirectional transformers for language
  understanding.
\newblock {\em arXiv preprint arXiv:1810.04805}, 2018.

\bibitem{ge2018deep}
Weifeng Ge.
\newblock Deep metric learning with hierarchical triplet loss.
\newblock In {\em ECCV}, 2018.

\bibitem{MarylandPolyvore}
Xintong Han, Zuxuan Wu, Yu-Gang Jiang, and Larry~S. Davis.
\newblock Learning fashion compatibility with bidirectional lstms.
\newblock In {\em ACM MM}, 2017.

\bibitem{ResNet}
Kaiming He, Xiangyu Zhang, Shaoqing Ren, and Jian Sun.
\newblock Deep residual learning for image recognition.
\newblock In {\em CVPR}, 2016.

\bibitem{kang2017visually}
Wang-Cheng Kang, Chen Fang, Zhaowen Wang, and Julian McAuley.
\newblock Visually-aware fashion recommendation and design with generative
  image models.
\newblock In {\em ICDM}, 2017.

\bibitem{kang2019complete}
Wang-Cheng Kang, Eric Kim, Jure Leskovec, Charles Rosenberg, and Julian
  McAuley.
\newblock Complete the look: Scene-based complementary product recommendation.
\newblock In {\em CVPR}, 2019.

\bibitem{GCN}
Thomas~N. Kipf and Max Welling.
\newblock Semi-supervised classification with graph convolution networks.
\newblock In {\em ICLR}, 2017.

\bibitem{li2017mining}
Yuncheng Li, Liangliang Cao, Jiang Zhu, and Jiebo Luo.
\newblock Mining fashion outfit composition using an end-to-end deep learning
  approach on set data.
\newblock {\em TMM}, 2017.

\bibitem{mcauley2015image}
Julian McAuley, Christopher Targett, Qinfeng Shi, and Anton Van Den~Hengel.
\newblock Image-based recommendations on styles and substitutes.
\newblock In {\em SIGIR}, 2015.

\bibitem{movshovitz2017no}
Yair Movshovitz-Attias, Alexander Toshev, Thomas~K Leung, Sergey Ioffe, and
  Saurabh Singh.
\newblock No fuss distance metric learning using proxies.
\newblock In {\em ICCV}, 2017.

\bibitem{schroff2015facenet}
Florian Schroff, Dmitry Kalenichenko, and James Philbin.
\newblock Facenet: A unified embedding for face recognition and clustering.
\newblock In {\em CVPR}, 2015.

\bibitem{shih2018compatibility}
Yong-Siang Shih, Kai-Yueh Chang, Hsuan-Tien Lin, and Min Sun.
\newblock Compatibility family learning for item recommendation and generation.
\newblock In {\em AAAI}, 2018.

\bibitem{SCE-net}
Reuben Tan, Mariya~I. Vasileva, Kate Saenko, and Bryan~A. Plummer.
\newblock Learning similarity conditions without explicit supervision.
\newblock In {\em ICCV}, 2019.

\bibitem{Type-aware}
Mariya~I. Vasileva, Bryan~A. Plummer, Krishna Dusad, Shreya Rajpal, Ranjitha
  Kumar, and David Forsyth.
\newblock Learning type-aware embeddings for fashion copatibility.
\newblock In {\em ECCV}, 2018.

\bibitem{CSN}
Andreas Veit, Serge Belongie, and Theofanis Karaletso.
\newblock Conditional similarity networks.
\newblock In {\em CVPR}, 2017.

\bibitem{SiameseNet}
Andreas Veit, Balazs Kovacs, Sean Bell, Julian McAuley, Kavita Bala, and Serge
  Belongie.
\newblock Learning visual clothing style with heterogeneous dyadic
  co-occurrences.
\newblock In {\em ICCV}, 2015.

\bibitem{wang2019multi}
Xun Wang, Xintong Han, Weilin Huang, Dengke Dong, and Matthew~R Scott.
\newblock Multi-similarity loss with general pair weighting for deep metric
  learning.
\newblock In {\em CVPR}, 2019.

\bibitem{weinberger2009distance}
Kilian~Q Weinberger and Lawrence~K Saul.
\newblock Distance metric learning for large margin nearest neighbor
  classification.
\newblock {\em JMLR}, 2009.

\bibitem{yu2019personalized}
Cong Yu, Yang Hu, Yan Chen, and Bing Zeng.
\newblock Personalized fashion design.
\newblock In {\em ICCV}, 2019.

\bibitem{zheng2019hardness}
Wenzhao Zheng, Zhaodong Chen, Jiwen Lu, and Jie Zhou.
\newblock Hardness-aware deep metric learning.
\newblock In {\em CVPR}, 2019.

\end{thebibliography}
}

\end{document}